\documentclass[runningheads]{llncs}

\usepackage{eccvabbrv}
\usepackage{dsfont}
\usepackage{graphicx}
\usepackage{booktabs}

\usepackage[accsupp]{axessibility}  

\usepackage{xcolor}
\definecolor{eccvblue}{RGB}{0, 102, 204}
\usepackage[pagebackref,breaklinks,colorlinks,citecolor=eccvblue]{hyperref}

\usepackage{orcidlink}

\begin{document}










\title{Adaptive Spectrum-Aware Feature Disentangled Network for Small Object Detection} 
\titlerunning{SFDNet}

\author{Yang Guo\inst{1,2}\orcidlink{0009-0000-0455-3217} \and
Zihan Yang\inst{3}\orcidlink{0009-0003-8818-5360} \and
Feifei Kou\inst{2} \and 
Yulan Hu\inst{4} \and
Ran Zhang\inst{5} \and
Siyuan Yao\inst{1}\thanks{Corresponding Author}\orcidlink{0000-0002-8479-5124}}

\authorrunning{Y.~Guo et al.}

\institute{
Shenzhen Campus of Sun Yat-sen University\\
\and
Beijing University of Posts and Telecommunications, Beijing, China\\ 
\and
Hangzhou International Innovation Institute, Beihang University, Hangzhou, China\\
\begin{center}
\inst{4} Renmin University of China \quad\quad
\inst{5} China CITIC Bank
\end{center}
\email{guoyang4409@gmail.com}\quad \email{zihanyang@buaa.edu.cn}\quad
\email{koufeifei000@bupt.edu.cn}\quad
\email{huyulan@ruc.edu.cn}\quad
\email{zhangran4@citicbank.com}\quad
\email{yaosiyuan04@gmail.com}}

\maketitle

\begin{abstract}
Small Object Detection (SOD) is a fundamental yet challenging problem in computer vision due to its limited spatial resolution and weak visual cues. Although recent approaches have achieved remarkable advances, the background distractors in different frequency spectra still degrade the performance. In this paper, we propose a novel small object detection framework termed \textbf{SFDNet}, which is capable of detecting small objects via efficient spectrum-aware feature disentanglement. Specifically, we propose an Adaptive Spectrum Disentanglement (ASD) module that decomposes backbone features into multiple complementary spectral components, aiming to construct discriminative object-relevant representations by discarding the background distractors for each component. Afterwards, to strengthen the semantic consistency of the similar objects in the same class, we propose a Class-Wise Prototype Distillation (CPD) procedure, which establishes class prototypes for the object instances and enforces the compact representation by efficient prototype distillation. Extensive experiments on multiple challenging benchmarks show that SFDNet outperforms existing state-of-the-art methods by a large margin. Our code is available at 
\url{https://github.com/ManOfStory/SFDNet}.
  \keywords{Small Object Detection \and Spectrum Disentanglement \and Prototype Distillation}
\end{abstract}

\section{Introduction}
\label{sec:intro}
Small Object Detection (SOD) is a critical research task aiming to accurately identify objects with limited size, which plays a vital role in a wide range of applications such as surveillance, drone-based inspection, and autonomous driving, etc. Compared to the classical object detection, the small objects are featured by fewer pixels and are prone to being confused with backgrounds, making it infeasible to adapt generic object detectors directly to the SOD task.

\begin{figure}[t]
\centering

\includegraphics[ width=1.0\linewidth]{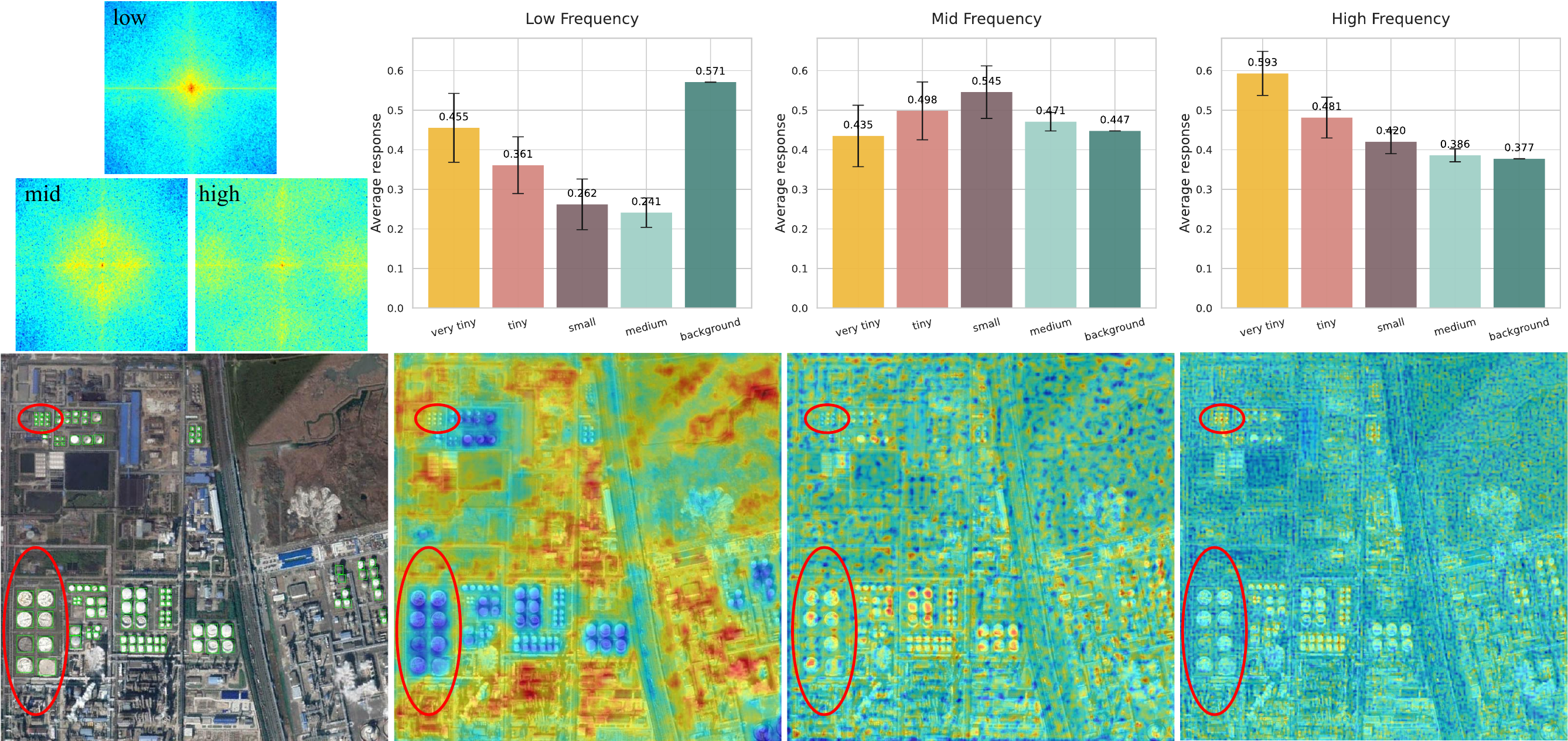}
\caption{The first row (left) illustrates the disentangled low, mid, and high frequency spectra, while the first row (right) reports the corresponding average response statistics within target regions at different scales. The second row presents the ground-truth annotations along with the heatmaps derived from low, mid, and high frequency features, respectively. Please zoom in for details. 
Regions highlighted by red circles emphasize salient differences.
}
\label{fig:motivation}
\end{figure}

From a technical perspective, existing small object detectors~\cite{xu2020gliding,yang2022querydet,li2022oriented,zhou2024kldet} primarily focus on learning semantically discriminative features to mitigate spatial information attenuation. Despite remarkable progress, these methods still struggle to reliably detect small objects in extremely complex scenarios. First, due to the limited spatial extent and weak visual cues of small objects, current detectors are easily confused by surrounding distractors. The most advanced approaches \cite{HS-FPN,sun2025set,li2026dyfclt} attempt to alleviate this issue by suppressing noise in specific frequency bands, thus the models are encouraged to pay more attention to the informative small-object features. However, these methods overlook the fact that noise is distributed across the entire spectrum, which is crucial for robust small object detection. As shown in Fig.~\ref{fig:motivation}, different frequency spectra exhibit distinct sensitivities to various scales of objects: the low-frequency spectrum tends to emphasize background structures and large-scale distractors, the mid-frequency spectrum demonstrates a wider range of response to the objects from tiny to medium-sized targets, while the high-frequency spectrum is particularly sensitive to extremely tiny objects. Due to the absence of comprehensive full-spectrum modeling, the performance of existing SOD methods\cite{CFINet,HS-FPN,10518058} is inherently constrained. Besides, many small objects share similar appearance semantics, while the intrinsic relationship information of these objects has not been sufficiently explored. Although some methods, e.g.,~\cite{wu2020self,kim2021robust,CFINet,yao2025unctrack}, attempt to enhance the appearance representation of small objects via super-resolution, feature mimic learning, or memory-based retrieval mechanisms, they still fail to effectively exploit the semantic similarity among objects. These approaches typically utilize the grouped samples with predefined metrics, e.g., IoU or cosine similarity, to construct instance-level relationships, but the generalization ability is still limited due to the lack of class-wise supervision.

To address these challenges, we propose SFDNet, which adaptively disentangles features into multiple complementary spectral components and enhances each component with tailored scanning strategies to better capture frequency-specific cues. By adaptively regulating the contributions of different spectrum components, SFDNet explicitly separates object-relevant information from background distractors within different spectra to facilitate accurate small object detection. 
In addition, we introduce a Class-Wise Prototype Distillation (CPD) procedure to model the shared appearance semantics among small objects within the same category. 
Through the distillation of class-specific semantic prototypes, CPD promotes compact and discriminative feature representations for each class. 
Extensive experiments on several widely used benchmarks demonstrate the effectiveness of the proposed method.
In summary, the main contributions of this paper are as follows:
\begin{itemize}
    \item We propose SFDNet, a novel small object detection framework that introduces an Adaptive Spectrum Disentanglement (ASD) module to adaptively disentangle features into multiple complementary spectral components, effectively mitigating background distractors.
    
    \item We introduce a Class-Wise Prototype Distillation (CPD) procedure, which performs category-aware prototype distillation to enforce compact representations, leading to more accurate and robust bounding box estimation.
    
    \item Extensive experiments on several popular benchmarks demonstrate the effectiveness and robustness of SFDNet in detecting extremely small objects under both densely and sparsely distributed scenarios.
\end{itemize}

\section{Related Work}
\subsection{Small Object Detection}
Recent advances in small object detection primarily focus on three aspects: data augmentation, label assignment, and feature enhancement. For data augmentation, ~\cite{kisantal2019augmentation,chen2019rrnet} augment small object instances via random copy-paste. DS-GAN~\cite{DS-GAN} utilizes a GAN-based generator to produce high-quality synthetic training samples. Fang et al.~\cite{fang2024data} propose a controllable diffusion-based augmentation method to generate high-quality training samples. For label assignment, NWD-RKA~\cite{xu2022detecting} and RFLA~\cite{xu2022rfla} leverage Wasserstein distance and KL divergence to mitigate the scale-variant IoU issue and increase positive samples for small objects in the training phase, while CFINet~\cite{CFINet} adopts a coarse-to-fine RPN with dynamic IoU thresholds. It also generates more effective positive samples across object scales. For feature enhancement, 
DN-FPN~\cite{10518058} employs contrastive learning to suppress noise in the top-down pathway of FPN by enhancing the discriminability of features at each level. 
HS-FPN~\cite{HS-FPN} first introduces a High-Frequency Perception (HFP) module to suppress low-frequency components that often dominate small object features. It also proposes a Spatial Dependency Perception (SDP) module to reduce the feature misalignment across FPN levels. However, its frequency filtering strategy relies on manually designed hyperparameters, which limit its ability to adaptively suppress background distractors in complex scenes.

Different from previous works, our method performs adaptive spectrum disentanglement by decomposing representations into different complementary spectral components, which explicitly separates object-relevant signals from background distractors, enabling accurate small object detection.

\subsection{Spectral-Aware Representation}
Spectral modeling has recently gained increasing attention in computer vision. Recent advances have incorporated spectral priors into convolutional architectures to enhance feature representation. For instance,
FADC~\cite{chen2024frequency} adaptively adjusts dilation rates based on local spectral characteristics, while FDConv~\cite{chen2025frequency} reformulates convolutional filters in the Fourier domain to enrich filter diversity and representation capacity.
Beyond convolutional adaptations, spectral decomposition has also been integrated into sequence modeling and robustness-oriented frameworks. TinyVim~\cite{TinyVim} introduces a Laplace-based mixer to decouple spectral components, strengthening visual representation learning within sequence modeling architectures. Bi et al.~\cite{Bi_2025_CVPR} leverage the Discrete Sine Transform to partition features into illumination-sensitive and illumination-insensitive spectral bands, thereby improving robustness under illumination variations. Furthermore, Wang et al.~\cite{wang2023generalized} employ spectral decomposition to disentangle domain-invariant and domain-specific representations, enhancing cross-domain generalization in UAV-based object detection.

In this study, we disentangle backbone features into multiple complementary spectral components and conduct spectrum-specific aggregation within each frequency spectrum. By adaptively fusing the resulting spectral representations, we obtain more discriminative features tailored for small object detection.

\subsection{State Space Model}
State Space Models (SSMs), originally developed in control theory, have recently been incorporated into deep learning to model sequential data by mapping a one-dimensional function or sequence to another sequence through an implicit latent state. However, classical SSMs are constrained by Linear Time Invariance (LTI), which imposes fundamental limitations on their capacity to model complex and input-dependent patterns. To address this issue, Mamba~\cite{gu2023mamba} introduces a Selective State Space Model that removes the LTI constraint while preserving linear-time efficiency.
Building upon Mamba, ~\cite{liu2024vmamba,zhu2024vision} adapts SSMs from one-dimensional sequences to two-dimensional visual data by employing multiple scanning routes to convert 2D feature maps into 1D sequences. MambaVision~\cite{hatamizadeh2025mambavision} further integrates Mamba with Vision Transformers (ViTs) by redesigning the Mamba formulation to enhance its effectiveness in modeling visual representations. More recently, Spatial-Mamba~\cite{xiao2025spatialmamba} observes that sequential scanning along any fixed direction inevitably disrupts spatial structure, and thus incorporates local spatial neighborhood features into each state to preserve the inherent 2D spatial relationships. In this work, we utilize SSMs in spectrum modeling and propose a multi-spectrum scan strategy.

\begin{figure*}[!t]
\centering
\includegraphics[width=\textwidth]{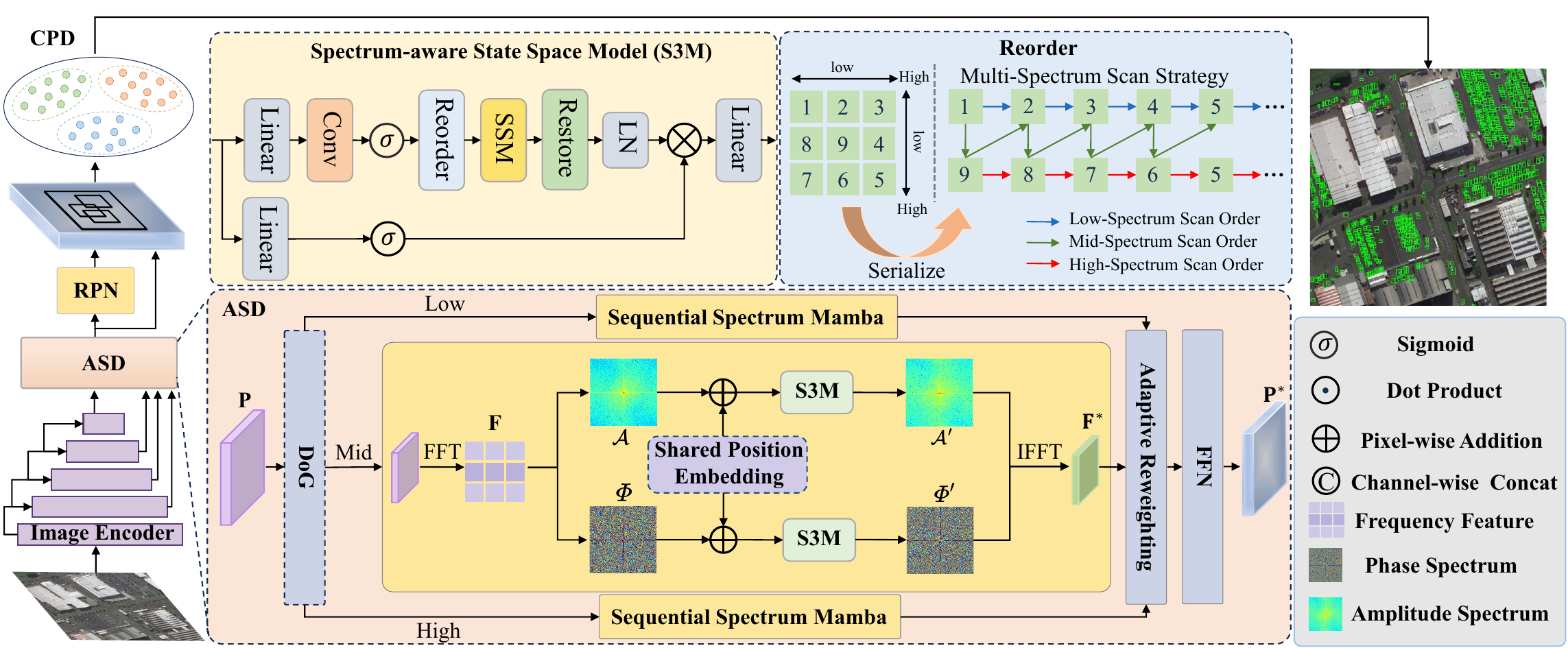}
\caption{The proposed architecture consists of two core components: the Adaptive Spectrum Disentanglement (ASD) module and the Class-Wise Prototype Distillation (CPD) procedure.  
The ASD module disentangles features into multi-spectrum components to perform full-spectrum suppression of background noise. 
The CPD procedure distills class-wise prototype representations, promoting compact and discriminative feature embeddings for each class.
}
\label{fig:framework}
\end{figure*}

\section{Methodology}
In this section, we present the overall architecture of SFDNet. As illustrated in Fig.~\ref{fig:framework}, the framework consists of two key components: an Adaptive Spectrum Disentanglement (ASD) module and a Class-Wise Prototype Distillation (CPD) procedure. 
The ASD module decomposes backbone features into multiple complementary spectral components and models the contextual dependencies of each spectrum under a spectrum-specific scanning strategy. 
The CPD procedure constructs class-wise prototypes and promotes compact and discriminative feature representations by exploiting intra-class similarity, thereby enhancing small object detection performance.

\subsection{Adaptive Spectrum Disentanglement}
Given an input image $\mathbf{I}_{t}$, we first feed it into a vision backbone to extract multi-scale feature maps $\{\mathbf{C}_2, \mathbf{C}_3, \mathbf{C}_4, \mathbf{C}_5\}$. These features are subsequently processed by a standard Feature Pyramid Network (FPN) to generate top-down aggregated multi-scale representations $\{\mathbf{P}_2, \mathbf{P}_3, \mathbf{P}_4, \mathbf{P}_5\}$. 
Without loss of generality, we denote an arbitrary pyramid level as $\mathbf{P}$. The aggregated feature $\mathbf{P}$ is then fed into the proposed Adaptive Spectrum Disentanglement (ASD) module for spectrum-aware processing.

Specifically, the ASD module first disentangles the feature $\mathbf{P}$ into low, mid, and high frequency components using the Difference of Gaussian (DoG) approach, formulated as follows:
\begin{equation}
\begin{aligned}
\mathbf{P}_{\text{low}} &= \mathcal{F}^{-1}\big[\mathcal{F}(\mathbf{P}) \cdot \hat{G}(\omega, k\sigma)\big],\\[1mm]
\mathbf{P}_{\text{mid}} &= \mathcal{F}^{-1}\big[\mathcal{F}(\mathbf{P}) \cdot (\hat{G}(\omega, \sigma) - \hat{G}(\omega, k\sigma))\big],\\[1mm]
\mathbf{P}_{\text{high}} &= \mathcal{F}^{-1}\big[\mathcal{F}(\mathbf{P}) \cdot (1 - \hat{G}(\omega, \sigma))\big],
\end{aligned}
\end{equation}
where $\hat{G}(\omega, \sigma)$ denotes the frequency-domain Gaussian filter with standard deviation $\sigma$, $\mathcal{F}$ and $\mathcal{F}^{-1}$ represent the Fourier transform and inverse Fourier transform, $\mathbf{P}_{\text{low}}$, $\mathbf{P}_{\text{mid}}$, and $\mathbf{P}_{\text{high}}$ correspond to the disentangled features of different frequency, respectively. 

The disentangled features are subsequently fed into the Sequential Spectrum Mamba module for spectrum-aware contextual modeling. Supposing $\hat{\mathbf{P}} \in \{\hat{\mathbf{P}}_{\text{low}}, \hat{\mathbf{P}}_{\text{mid}}, \hat{\mathbf{P}}_{\text{high}}\}$ to uniformly denote the frequency-specific features of each spectrum. Specifically, $\hat{\mathbf{P}}$ is first transformed into the frequency domain via the Fast Fourier Transform (FFT), yielding the corresponding frequency-domain representation $\mathbf{F}$. The FFT is formulated as follows:
\begin{equation}\label{eq:P_FFT} 
\mathbf{F}(u,v) = \sum_{x=0}^{H-1} \sum_{y=0}^{W-1} \hat{\mathbf{P}}(x,y) \cdot \exp\left(-j 2\pi \left( \frac{ux}{H} + \frac{vy}{W} \right) \right),
\end{equation}
where $\hat{\mathbf{P}}(x, y)$ denotes the spatial domain feature at location $(x, y)$, $\mathbf{F}(u, v)$ is its corresponding frequency representation at coordinate $(u,v)$. 
Afterwards, the frequency features $\mathbf{F}$ are then decomposed into the amplitude spectrum $\mathcal{A}$ and phase spectrum $\Phi$, which can be computed as follows:
\begin{equation}
\begin{aligned}
\mathcal{A}(u, v) &= \sqrt{\operatorname{Re}(\mathbf{F}(u, v))^2 + \operatorname{Im}(\mathbf{F}(u, v))^2}, \\
\Phi(u, &v) = \arctan\left( \frac{\operatorname{Im}(\mathbf{F}(u, v))}{\operatorname{Re}(\mathbf{F}(u, v))} \right),
\end{aligned}
\label{eq:amp_phase}
\end{equation}
where $\operatorname{Re}(\cdot)$ and $\operatorname{Im}(\cdot)$ denote the real and imaginary parts of the complex-valued feature, respectively. The amplitude spectrum $\mathcal{A}$ and phase spectrum $\Phi$ are then added with a shared learnable positional embedding, respectively. The resulting features are then fed into Sequential State Space Models (S3M), which aggregate information within different spectra to capture long-range spectral dependencies, as detailed in Sec.~\ref{S3M}. The complete process is formulated as follows:
\begin{equation}
\begin{aligned}
\mathcal{A}^{\prime}(u, v) &= \mathrm{S3M}(\mathcal{A}(u, v)+\mathrm{PE}(u,v)), \\
\Phi^{\prime}(u, v) &= \mathrm{S3M}(\Phi(u, v)+\mathrm{PE}(u,v)), \\
\end{aligned}
\label{eq:reconstruction}
\end{equation}
where $\mathcal{A}^{\prime}\in \mathbb{R}^{C\times H\times W}$ and $\Phi^{\prime}\in \mathbb{R}^{C\times H\times W}$ denote the modified amplitude and phase spectra, respectively.  
The refined amplitude and phase spectra are then transformed back into the spatial domain via the inverse Fast Fourier Transform (IFFT), yielding the enhanced feature representation $\mathbf{F}^*$.

To effectively integrate information from different spectra, the spectrum-aware disentangled features are adaptively reweighted according to their learned importance vectors $\boldsymbol{\alpha}_\text{L}$, $\boldsymbol{\alpha}_\text{M}$, and $\boldsymbol{\alpha}_\text{H} \in \mathbb{R}^C$. The reweighted features are subsequently fused through a Feed-Forward Network (FFN) to produce the aggregated representation $\mathbf{P}^{*}$.
\begin{equation}
\mathbf{P}^{*} = \mathrm{FFN}\big([\boldsymbol{\alpha}_\text{L},\boldsymbol{\alpha}_\text{M},\boldsymbol{\alpha}_\text{H}][\mathbf{F}_{\text{low}}^{*}, \mathbf{F}_{\text{mid}}^{*}, \mathbf{F}_{\text{high}}^{*}]^\text{T}\big),
\end{equation}
The fused representation $\mathbf{P}^{*}$ is subsequently utilized for downstream detection, serving as a spectrum-adaptively enhanced feature embedding for small object detection.

\begin{figure*}[!t]
\centering
\includegraphics[width=\textwidth]{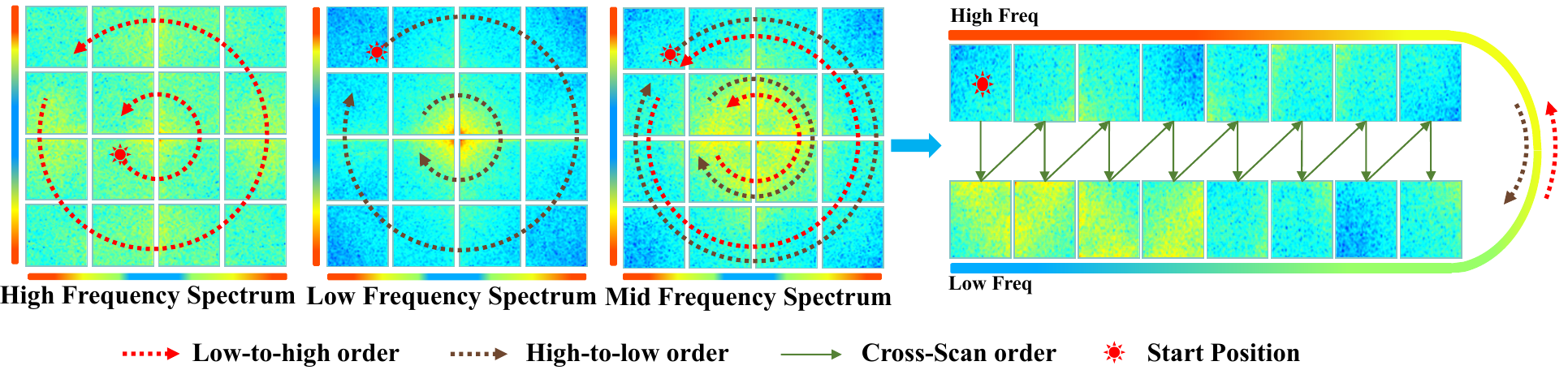}
\caption{Illustration of the Multi-Spectrum Scan Strategy. The high and low frequency spectra follow low-to-high and high-to-low scanning orders, respectively, whereas the mid-frequency spectrum adopts an alternating cross-scan strategy.
}
\label{fig:scanning_strategy}
\end{figure*}

\subsection{Spectrum-aware State Space Model (S3M)}
\label{S3M}
As illustrated in Fig.~\ref{fig:framework}, we propose a Multi-Spectrum Scan Strategy that serializes each spectrum map into a one-dimensional sequence, enabling long-range contextual modeling over the entire spectrum. Since State Space Models (SSMs) process data sequentially by propagating historical information through hidden states, the state vector is inherently dominated by nearby elements and less affected by the remote ones. When modeling along the spectral dimension, this order-sensitive processing induces spectral biases that depend on the chosen scanning order.

To effectively aggregate information emphasized by different spectra, we leverage this property of SSMs to design spectrum-specific scanning strategies tailored to the characteristics of each frequency band. As shown in Fig.~\ref {fig:scanning_strategy}, for the high-frequency spectrum, we adopt a low-to-high scanning order that unfolds the spectrum in a spiral manner from the spectral center outward along increasing radii, biasing the aggregation toward high-frequency components. Conversely, for the low-frequency spectrum, we follow a high-to-low scanning order, such that low-frequency components dominate the resulting representation. For the mid-frequency spectrum, we employ an alternating cross-scan strategy that interleaves inner and outer spectral regions, mixing low and high frequency signals during serialization to mitigate the inherent inductive bias of SSMs and promote balanced contextual modeling across the spectrum. 

The reordered sequences are subsequently fed into the SSM to capture long-range dependencies across frequency signals. Following the discretized formulation of Mamba, the state transition is defined as
\begin{equation}
\begin{aligned}
\bar{\mathbf{A}}_t &= \exp(\Delta_t \mathbf{A}) 
\\ 
\bar{\mathbf{B}}_t &= (\Delta_t \mathbf{A})^{-1}(\exp(\Delta_t \mathbf{A}) - \mathbf{I})\Delta_t \mathbf{B_t}  
\\
\mathbf{h}_t &= \bar{\mathbf{A}}_t \mathbf{h}_{t-1} + \bar{\mathbf{B}}_t \mathbf{x}_t, \quad
\mathbf{y}_t = \mathbf{C_t} \mathbf{h}_t,
\end{aligned}
\end{equation}
where $\mathbf{A}$ denotes a learnable input-invariant state transition embedding, while $\mathbf{B}_t$, $\mathbf{C}_t$, and $\Delta_t$ are input-variant parameters dynamically generated from the input $\mathbf{x}_t$. The hidden state $\mathbf{h}_t$ represents the latent state at step $t$, and $\bar{\mathbf{A}}_t$ and $\bar{\mathbf{B}}_t$ are the corresponding discretized state transition matrices that enable selective and input-dependent state evolution in the SSM.

Finally, the aggregated spectrum sequences are restored to their original spatial layout via the inverse permutation, ensuring that the underlying frequency basis remains invariant before and after the transformation.

\subsection{Class-Wise Prototype Distillation}
\begin{figure}[!t]
\centering
\includegraphics[width=\textwidth]{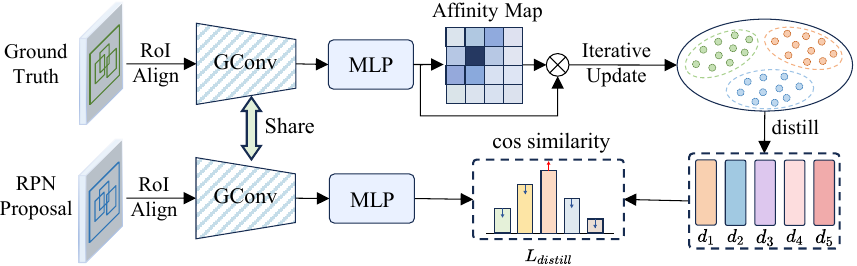}
\caption{
Overview of the Class-Wise Prototype Distillation (CPD) procedure. 
CPD distills class-specific prototypes from ground-truth RoI features, which are then used to guide proposal features via contrastive supervision, enhancing semantic consistency within each category.
}
\label{fig:cpd_procedure}
\end{figure}

To better capture the relational semantics among intra-class instances, as shown in Fig.\ref{fig:cpd_procedure}, we introduce a Class-Wise Prototype Distillation (CPD) procedure that explicitly constructs category-level prototypes from ground-truth annotations. By grouping the prototype representation, the hard samples related to the same class can be mutually enhanced, leading to more discriminative and category-consistent representations.

During training, we extract RoI features from ground-truth bounding boxes, collapse the spatial dimensions using a Global Convolution layer, and project them through an MLP to obtain embedding vectors $\mathbf{Z} = [\mathbf{z}_1, \dots, \mathbf{z}_N]^\top$. Meanwhile, we establish a class-wise prototype bank $\{\mathbf{d}_c \in \mathbb{R}^C\}_{c=1}^{N_C}$ in the embedding space. At initialization, each prototype $\mathbf{d}_c^{(0)}$ is computed as the empirical mean of the ground-truth RoI embeddings belonging to class $c$. To further refine these prototypes and capture intra-class variations, we formulate prototype distillation as a class-conditional, affinity-guided refinement procedure inspired by the hierarchical affinity modeling in \cite{yao2024hierarchical} and the expectation-maximization paradigm. For each class $c$, the class-aware affinity and the refined prototype are defined as:
\begin{equation}
\mathbf{A}_{i,c}
=
\frac{
\exp\!\left(\kappa \mathbf{z}_i^\top \mathbf{d}_c\right)
\, \mathds{1}(i \in \Omega^{gt}_c)
}
{
\sum_{j=1}^{N}
\exp\!\left(\kappa \mathbf{z}_j^\top \mathbf{d}_c\right)
\, \mathds{1}(j \in \Omega^{gt}_c)
},
\quad
\mathbf{d}_c^{*}
=
\sum_{i=1}^{N}
\mathbf{A}_{i,c} \mathbf{z}_i,
\end{equation}
where $\mathds{1}(\cdot)$ denotes the indicator function, $\Omega^{gt}_c$ is the index set of ground-truth instances of class $c$ in the current mini-batch, and $\kappa$ controls the concentration of the similarity distribution.
Afterwards, the prototype bank is updated via a momentum scheme across minibatches to preserve global stability:
\begin{equation}
\mathbf{d}_c 
\leftarrow 
\alpha \mathbf{d}_c 
+
(1-\alpha)\mathbf{d}_c^{*}.
\end{equation}
With the refined prototype bank, we further impose contrastive alignment on proposal features. Given object proposals generated by the RPN, we extract RoI features using RoI Align from $\mathbf{P}^*$ and project them into the same embedding space, yielding instance representations $\mathbf{B} \in \mathbb{R}^{N \times C}$. To align these proposals with their corresponding class prototypes, we adopt an InfoNCE-based contrastive objective. For each embedding $\mathbf{b}_i \in \mathbf{B}$ belonging to class $c$, its similarity to prototype $\mathbf{d}_c$ is maximized against other prototypes $\{\mathbf{d}_k\}_{k=1}^{N_c}$. The loss is formulated as:
\begin{equation}
\mathcal{L}_{\text{distill}} = - \frac{1}{N} \sum_{c=1}^{N_c} \sum_{i \in \Omega_c} 
\log \frac{\exp \big(\mathrm{cos}(\mathbf{b}_i, \mathbf{d}_c) / \tau \big)}
{\sum_{k=1}^{N_c} \exp \big(\mathrm{cos}(\mathbf{b}_i, \mathbf{d}_k) / \tau \big)},
\label{eq:info_nce_proto}
\end{equation}
where $N_c$ denotes the number of categories, $\Omega_c$ is the index set of RoIs belonging to class $c$, $\mathrm{cos}(\cdot,\cdot)$ denotes cosine similarity, and $\tau$ is a temperature parameter.
Finally, the overall loss function is defined as:
\begin{equation}
\begin{aligned}
    \mathcal{L} = \mathcal{L}&_{\text{IoU}} + \mathcal{L}_{\text{cls}} + \gamma \mathcal{L_\text{distill}},
\end{aligned}
\end{equation}
where $\mathcal{L}_{\text{IoU}}$ denotes the IoU loss. $\mathcal{L}_{\text{cls}}$ is the cross-entropy classification loss for candidate proposals. $\mathcal{L}_{distill}$ is the distillation loss. The coefficient $\gamma$ is a hyperparameter that balances the loss contribution.

\setlength{\tabcolsep}{1mm}
\begin{table*}[!t]
\caption{Comparison with state-of-the-art detection methods on AI-TOD test set. All metrics follow the COCO-style evaluation protocol, including overall AP and performance across different scene types. \textbf{Bold} numbers indicate the best results. $^{\dagger}$ denotes results trained using the official implementation.}
\centering
\resizebox{0.85\textwidth}{!}{
\begin{tabular}{l|c|ccccccc}
\hline
Method & Source & AP & AP$_{50}$ & AP$_{75}$ & AP$_{\text{vt}}$ & AP$_{\text{t}}$ & AP$_{\text{s}}$ & AP$_{\text{m}}$ \\ \hline
\multicolumn{9}{l}{\textbf{End-to-End Detectors}} \\\hline
DAB-DETR\cite{DBLP:conf/iclr/LiuLZYQSZZ22} & ICLR2022 & 4.9 & 16.0 & 1.7 & 1.7 & 3.6 & 7.0 & 18.0 \\
DAB-Deformable-DETR\cite{DBLP:conf/iclr/LiuLZYQSZZ22} & ICLR2022 & 16.5 & 42.6 & 9.9 & 7.9 & 15.2 & 23.8 & 31.9 \\
DINO-Deformable-DETR\cite{DBLP:conf/iclr/0097LL000NS23} & ICLR2023 & 23.2 & 56.6 & 15.4 & 9.9 & 23.1 & 29.3 & 37.6 \\
DINO-5scale w/SET \cite{sun2025set} & CVPR2025 & 26.6 & 57.1 & 20.8 & 13.2 & 27.1 & 31.5 & -- \\\hline
\multicolumn{9}{l}{\textbf{Multi-Stage Detectors}} \\\hline
Faster R-CNN \cite{ren2016faster} & PAMI2017 & 11.1 & 26.3 & 7.6 & 0.0 & 7.2 & 23.3 & 33.6 \\
Cascade R-CNN \cite{cai2018cascade} & CVPR2018 & 13.8 & 30.8 & 10.5 & 0.0 & 10.5 & 25.5 & 36.6 \\
DetectorRS\cite{qiao2021detectors} & CVPR2021 & 14.8 & 32.8 & 11.4 & 0.0 & 10.8 & 28.3 & 38.0 \\
QueryDet\cite{yang2022querydet} & CVPR2022  & 12.2 & 29.3 & 7.9 & 2.4 & 10.5 & 18.5 & 26.3 \\
CFINet$^{\dagger}$\cite{CFINet} & ICCV2023 & 24.7 & 53.9 & 18.6 & 11.7 & 26.4 & 28.1 & 32.2 \\
RFLA\cite{xu2022rfla} & ECCV2022 & 24.8 & 55.2 & 18.5 & 9.3 & 24.8 & 30.3 & 38.2 \\
DNTR\cite{10518058} & TGRS2024 & 26.2 & 56.7 & 20.2 & 12.8 & 26.4 & 31.0 & 37.0 \\
SimD\cite{shi2024similarity} & IROS2024 & 26.6 & 55.9 & 21.2 & 13.4 & 27.5 & 30.9 & 37.8 \\
DetectorRS w/FIDP \cite{bian2025feature} & CVPR2025 & 24.3 & 54.4 & 18.3 & 8.5 & 24.9 & 29.8 & -- \\
HS-FPN\cite{HS-FPN} & AAAI2025 & 25.1 & 55.7 & 19.1 & 12.1 & 25.3 & 29.9  & 36.9 \\ 
\hline
SFDNet & -- & 29.0 & 57.9 & 23.5 & 14.4 & 29.0 & 33.4 & 40.8
\\
{SFDNet}$^*$ & -- & \textbf{31.7} & \textbf{64.9} & \textbf{25.6} & \textbf{17.6} & \textbf{32.8} & \textbf{36.1} & \textbf{42.5}
\\ 

\hline
\end{tabular}
}
\label{tab:aitod_comparison}
\end{table*}

\section{Experiment}
In this section, we first provide the implementation details of our method and comparisons with other state-of-the-art detectors on three popular benchmarks: \textbf{AI-TOD}~\cite{wang2021tiny}, \textbf{SODA-D}~\cite{cheng2023towards} and \textbf{SODA-A}~\cite{cheng2023towards}. Next, we conduct ablation studies to evaluate the effectiveness of our approach. Finally, we present qualitative visualizations to analyze the impact on feature representations.

\subsection{Implementation Details}
All experiments are conducted on a single NVIDIA A100 GPU. For AI-TOD, input patches are fixed to a size of $800 \times 800$, whereas for SODA-D and SODA-A, the patches are resized to $1200 \times 1200$. The hyperparameters $\tau$ and $\gamma$ are set to 0.07 and 0.1. The DoG parameters are set as $\sigma = 1.0$ and $k = 1.414$. The model is trained for 36 epochs on AI-TOD and 12 epochs on SODA-D and SODA-A. Based on different backbones, we propose two variants of SFDNet: SFDNet and SFDNet$^*$. SFDNet adopts a ResNet-50\cite{he2016deep} backbone, while SFDNet$^*$ employs a Spatial-Mamba\cite{xiao2025spatialmamba} backbone.

\subsection{Comparison with state-of-the-art}
\noindent\textbf{AI-TOD.}
The AI-TOD dataset contains 700,621 instances in 8 categories in 28,036 aerial images. Compared to existing object detection datasets in aerial images, the mean size of objects in AI-TOD is about 12.8 pixels, which is much smaller than others.
We report the results of our proposed SFDNet on AI-TOD test set. As shown in Table \ref{tab:aitod_comparison}, SFDNet$^*$ achieves the best performance across all metrics, surpassing other state-of-the-art methods by a large margin.  Compared with HS-FPN \cite{HS-FPN}, it improves (AP, AP$_{50}$, AP$_{75}$, AP$_{\text{vt}}$, AP$_{\text{t}}$, AP$_{\text{s}}$, AP$_{\text{m}}$) by (6.6\%, 9.2\%, 6.5\%, 5.5\%, 7.5\%, 6.2\%, 5.6\%), respectively. Besides, SFDNet also achieves state-of-the-art performance.

\setlength{\tabcolsep}{1mm} 
\begin{table*}[!t]
\caption{Comparison with oriented object detection methods on SODA-A test set. Metrics follow COCO-style evaluation. \textbf{Bold} numbers indicate the best results. $^{\dagger}$ denotes results trained using the official implementation.}
\centering
\renewcommand{\arraystretch}{1.05}
\resizebox{0.85\textwidth}{!}{
\begin{tabular}{l|c|ccccccc}
\hline
Method & Source & AP & AP$_{50}$ & AP$_{75}$ & AP$_\text{es}$ & AP$_\text{rs}$ & AP$_\text{gs}$ & AP$_{N}$ \\
\hline
Rotated Faster R-CNN\cite{ren2016faster}  & PAMI2017 & 32.5 & 70.1 & 24.3 & 11.9 & 27.3 & 42.2 & 34.4 \\
Rotated RetinaNet\cite{lin2017focal}  & ICCV2017 & 26.8 & 63.4 & 16.2 & 9.1  & 22.0 & 35.4 & 28.2 \\
Gliding Vertex\cite{xu2020gliding}   & PAMI2020 & 31.7 & 70.8 & 22.6 & 11.7 & 27.0 & 41.1 & 33.8 \\
Oriented R-CNN\cite{xie2021oriented} & ICCV2021 & 34.4 & 70.7 & 28.6 & 12.5 & 28.6 & 44.5 & 36.7 \\
S$^2$A-Net\cite{han2021align}  &TGRS2021 & 28.3 & 69.6 & 13.1 & 10.2 & 22.8 & 35.8 & 29.5 \\
DODet\cite{cheng2022dual} & TGRS2022 & 31.6 & 68.1 & 23.4 & 11.3 & 26.3 & 41.0 & 33.5 \\
Oriented RepPoints\cite{li2022oriented} & CVPR2022 & 26.3 & 58.8 & 19.0 & 9.4  & 22.6 & 32.4 & 28.5 \\
DHRec\cite{nie2022multi} & PAMI2022 & 30.1 & 68.8 & 19.8 & 10.6 & 24.6 & 40.3 & 34.6 \\
CFINet\cite{CFINet} &ICCV2023 & 34.4 & 73.1 & 26.1 & 13.5 & 29.3 & 44.0 & 35.9 \\
DecoupleNet$^{\dagger}$\cite{lu2024decouple} & TGRS2024 & 36.6 & 71.3 & 33.3 & 12.2 & 31.0 & 47.7 & 40.2 \\ 
LEGNet$^{\dagger}$\cite{lu2025legnet} & ICCVW2025 & 29.6 & 58.7 & 26.4 & 10.4 & 26.0 & 39.6 & 32.0 \\
GauCho$^{\dagger}$ \cite{Marques_2025_CVPR} & CVPR2025 & 33.2 & 70.1 & 25.0 & 9.9 & 27.8 & 44.9 & 36.4 \\
UGS\cite{sun2025uncertainty} & ICCV2025 & 36.0 & 73.1 & 30.3 & 13.7 & 30.2 & 47.8 & 38.1 \\
DCFL\cite{xu2025oriented} & PAMI2026 & 36.6 & 72.6 & 32.4 & 13.9 & 30.3 & 47.4 & 41.2 \\
Unc-SOD\cite{11361337} & TIP2026 & 34.8 & 73.6 & 26.4 & 13.8 & 29.7 & 44.7 & 36.5 \\
\hline
SFDNet & -- & 37.8 & 73.1 & 34.5 & 13.1 & 32.1 & 49.4 & 42.9 \\
SFDNet$^*$ & -- & \textbf{39.2} & \textbf{75.0} & \textbf{36.6} & \textbf{15.5} & \textbf{33.7} & \textbf{50.4} & \textbf{45.6} \\
\hline
\end{tabular}
}
\label{tab:sodaa_comparison}
\end{table*}

\noindent\textbf{SODA-A.}
The SODA-A dataset consists of 2,513 aerial images with 872,069 oriented bounding box annotations, focusing on small object detection. SODA-A has an average resolution of 4761 × 2777 pixels and contains approximately 347 instances per image, presenting a highly dense object distribution that challenges existing detection models in clustered scenes. Table~\ref{tab:sodaa_comparison} summarizes the performance on SODA-A benchmark. SFDNet$^*$ achieves state-of-the-art results with (39.2\%, 75.0\%, 36.6\%, 15.5\%, 33.7\%, 50.4\%, 45.6\%) on (AP, AP$_{50}$, AP$_{75}$, AP$_\text{es}$ AP$_\text{rs}$, AP$_\text{gs}$, AP$_\textbf{N}$), surpassing state-of-the-art GauCho~\cite{Marques_2025_CVPR} by (6.0\%, 4.9\%, 11.6\%, 5.6\%, 5.9\%, 5.5\%, 9.2\%) across all metrics.

\noindent\textbf{SODA-D.} The SODA-D dataset contains 24,828 high-resolution images with 278,433 annotated instances spanning 9 categories. It exhibits rich diversity in terms of locations, weather conditions, period, camera viewpoints, and traffic scenarios. With an average image resolution of 3407 × 2470, the dataset is particularly well-suited for detecting tiny and small objects in complex environments.
We report the performance of SFDNet on SODA-D test set. As summarized in Table~\ref{tab:sodad_comparison}, our SFDNet$^*$ achieves state-of-the-art performance across all metrics, with an average AP of 34.2\%. 
Compared with HS-FPN\cite{HS-FPN}, SFDNet$^*$ yields consistent improvements of (4.6\%, 7.2\%, 4.6\%, 3.1\%, 4.3\%, 5.3\%, 5.1\%) in terms of (AP, AP$_{50}$, AP$_{75}$, AP$_\text{es}$, AP$_\text{rs}$, AP$_\text{gs}$, AP$_\text{N}$).

\begin{table*}[!t]
\caption{Comparison with state-of-the-art detection approaches on SODA-D test set. All metrics follow COCO-style evaluation, including overall AP and performance across different scene types. \textbf{Bold} numbers indicate the best results. $^{\dagger}$ denotes results trained using the official implementation.}
\centering

\resizebox{0.85\textwidth}{!}{\begin{tabular}{l|c|ccccccc}
\hline
Method & Source & AP & AP$_{50}$ & AP$_{75}$ & AP$_{\text{es}}$ & AP$_{\text{rs}}$ & AP$_{\text{gs}}$ & AP$_N$ \\
\hline
Faster R-CNN\cite{ren2016faster} & PAMI2017 & 28.9 & 59.7 & 24.2 & 13.9 & 25.6 & 34.3 & 43.2 \\
RetinaNet\cite{lin2017focal}& ICCV2017 & 28.2 & 57.6 & 23.7 & 11.9 & 25.2 & 34.1 & 44.2 \\
CornerNet\cite{law2018cornernet} & ECCV2018 & 24.6 & 49.5 & 21.7 & 6.5  & 20.5 & 32.2 & 43.8 \\
FCOS\cite{tian2019fcos} & ICCV2019 & 23.9 & 49.5 & 19.9 & 6.9  & 19.4 & 30.9 & 40.9 \\
RepPoints\cite{yang2019reppoints} & ICCV2019 & 28.0 & 55.6 & 24.7 & 10.1 & 23.8 & 35.1 & 45.3 \\
ATSS\cite{zhang2020bridging} & CVPR2020 & 26.8 & 55.6 & 22.1 & 11.7 & 23.9 & 32.2 & 41.3 \\
Cascade RPN\cite{vu2019cascade} & NIPS2019 & 29.1 & 56.5 & 25.9 & 12.5 & 25.5 & 35.4 & 44.7 \\
Deformable-DETR\cite{zhu2021deformable} & ICLR2021 & 19.2 & 44.8 & 13.7 & 6.3  & 15.4 & 24.9 & 34.2 \\
Sparse R-CNN\cite{sun2021sparse} & CVPR2021 & 24.2 & 50.3 & 20.3 & 8.8  & 20.4 & 30.2 & 39.4 \\
DyHead\cite{dai2021dynamic} & CVPR2021 & 27.5 & 56.1 & 23.2 & 12.4 & 24.4 & 33.0 & 41.9 \\
RFLA\cite{xu2022rfla} & ECCV2022 & 29.7 & 60.2 & 25.2 & 13.2 & 26.9 & 35.4 & 44.6 \\
CFINet\cite{CFINet} & ICCV2023 & 30.7 & 60.8 & 26.7 & 14.7 & 27.8 & 36.4 & 44.6 \\ 
DNTR$^{\dagger}$\cite{10518058} & TGRS2024 & 29.6 & 57.8 & 26.5 & 13.1 & 26.7 & 35.5 & 43.4 \\
CFPT$^{\dagger}$\cite{11010850} & TGRS2025 & 27.5 & 54.5 & 23.8 & 7.1 & 22.4 & 36.2 & 45.9 \\
HS-FPN$^{\dagger}$\cite{HS-FPN} & AAAI2025 & 29.6 & 56.8 & 26.7 & 13.6 & 26.4 & 35.3 & 45.3 \\
Unc-SOD\cite{11361337} & TIP2026 & 31.0 & 60.8 & 27.1 & 14.9 & 27.6 & 36.9 & 45.8 \\
\hline
SFDNet & -- & 31.3 & 62.1 & 26.8 & 15.1 & 27.8 & 37.3 & 46.2 \\
SFDNet$^*$ & --  & \textbf{34.2} & \textbf{64.0} & \textbf{31.3} & \textbf{16.7} & \textbf{30.7} & \textbf{40.6} & \textbf{50.4} \\
\hline
\end{tabular}
}
\label{tab:sodad_comparison}
\end{table*}

\subsection{Ablation Study}
\noindent\textbf{Ablation of Different Variations.} To evaluate the effectiveness of each component, we selectively ablate ASD and CPD from SFDNet. 
As shown in Table~\ref{tab:ablation_variants}, the baseline detector without ASD and CPD achieves (24.8\%, 9.3\%, 24.8\%, 30.3\%, 38.2\%) in terms of (AP, AP$_\text{vt}$, AP$_\text{t}$, AP$_\text{s}$, AP$_\text{m}$), respectively. When ASD is introduced, the performance improves to (28.6\%, 13.8\%, 28.8\%, 33.2\%, 39.2\%), demonstrating its capability to effectively suppress noise across the full spectrum. Additionally, incorporating CPD into the baseline also leads to performance gains, achieving (27.2\%, 12.0\%, 27.6\%, 32.2\%, 38.3\%) for (AP, AP$_\text{vt}$, AP$_\text{t}$, AP$_\text{s}$, AP$_\text{m}$), respectively. Furthermore, combining both ASD and CPD yields the best performance, with improvements of (4.2\%, 5.1\%, 4.2\%, 3.1\%, 2.6\%) over the baseline. These results validate the effectiveness of the proposed ASD module and CPD procedure.

\noindent\textbf{Ablation Study of the Scanning Mechanism.} To validate the effectiveness of our Multi-Spectrum Scan Strategy, we conduct ablation experiments on different scanning mechanisms. Specifically, we compare several commonly used scanning methods, including sweeping, clockwise, and neighbor-based orders. As shown in Tab.~\ref{tab:ablation_scan_mechanism}, replacing our proposed scan strategy with any of these alternatives leads to performance degradation, demonstrating the effectiveness of the Multi-Spectrum Scan Strategy.

\begin{table}[!t]
\centering
\footnotesize
\renewcommand{\arraystretch}{1.0}
\setlength{\tabcolsep}{2.8pt}

\begin{minipage}[t]{0.48\linewidth}
\centering
\caption{Ablation studies of the different variants in SFDNet on AI-TOD dataset. The bold number indicates the best result obtained in the experiment.}
\label{tab:ablation_variants}
\begin{tabular}{cc|ccccc}
\hline
ASD & CPD & AP & AP$_\text{vt}$ & AP$_\text{t}$ & AP$_\text{s}$ & AP$_\text{m}$\\
\hline
$\times$ & $\times$ & 24.8 & 9.3 & 24.8 & 30.3 & 38.2\\
$\checkmark$ & $\times$ & 28.6 & 13.8 & 28.8 & 33.2 & 39.2\\
$\times$ & $\checkmark$ & 27.2 & 12.0 & 27.6 & 32.2 & 38.3\\
$\checkmark$ & $\checkmark$ & \textbf{29.0} & \textbf{14.4} & \textbf{29.0} & \textbf{33.4} & \textbf{40.8}\\
\hline
\end{tabular}
\end{minipage}
\hfill
\begin{minipage}[t]{0.48\linewidth}
\centering
\caption{Ablation study of the scanning mechanism in SFDNet on AI-TOD dataset. The bold number indicates the best result obtained in the experiment.}
\label{tab:ablation_scan_mechanism}
\begin{tabular}{c|ccccc}
\hline
Setting & AP & AP$_\text{vt}$ & AP$_\text{t}$ & AP$_\text{s}$ & AP$_\text{m}$\\
\hline
Sweep.~\cite{liu2024vmamba} & 26.7 & 13.6 & 27.1 & 30.6 & 36.9\\
Clock.~\cite{li2025mamba} & 27.0 & 13.1 & 27.4 & 31.2 & 36.9\\
Neigh.~\cite{xiao2025spatialmamba} & 26.2 & 12.3 & 26.6 & 29.7 & 38.5\\
Ours & \textbf{29.0} & \textbf{14.4} & \textbf{29.0} & \textbf{33.4} & \textbf{40.8}\\
\hline
\end{tabular}
\end{minipage}


\end{table}

\noindent\textbf{Ablation Study of Different Spectra.} To investigate the impact of different spectra, we conduct ablation experiments by selectively discarding each spectrum feature. As shown in Tab.~\ref{tab:Spectrum_impact}, removing any single spectrum component results in performance degradation. Specifically, the removal of the low-frequency spectrum leads to decreases of (3.0\%, 4.6\%, 4.0\%, 4.2\%) in terms of (AP$_\text{vt}$, AP$_\text{t}$, AP$_\text{s}$, AP$_\text{m}$), respectively. The removal of mid and high frequency components leads to a more pronounced performance degradation, demonstrating their stronger relevance to small objects.
These results demonstrate the critical importance of preserving full-spectrum information for small object detection.

\noindent\textbf{Ablation Study of Spectrum Aggregation Mechanism.} 
To evaluate the effectiveness of our proposed spectrum aggregation mechanism, e.g., S3M, we conduct ablation experiments by replacing S3M with alternative mechanisms. As shown in Tab.~\ref{tab:spectrum_aggregation}, substituting S3M with Attention, VMamba, or Convolution consistently results in performance degradation. These findings highlight the critical role of S3M in spectrum aggregation for small object detection.

\begin{figure*}[!t]

\centering
\caption{Per Class Performance on SODA-D and AI-TOD. Please zoom in for details.}
\label{fig:visualization_radar}
\includegraphics[width=\linewidth]{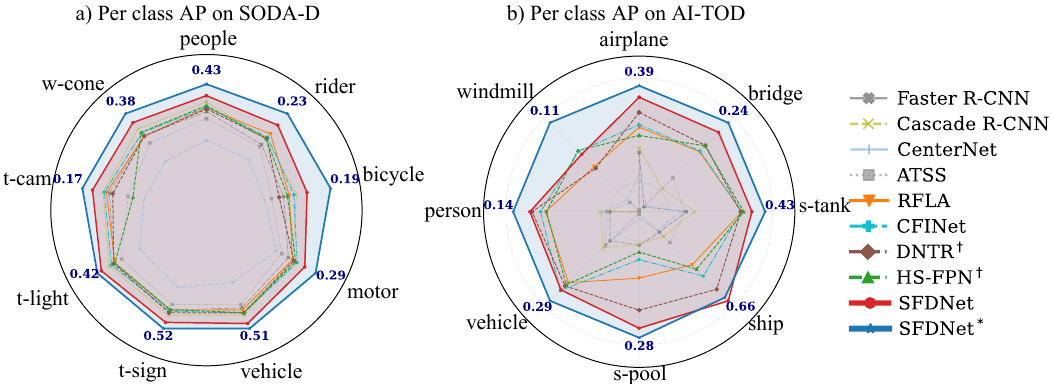}
\end{figure*}

\begin{table}[!t]
\centering
\footnotesize
\renewcommand{\arraystretch}{1.0}
\setlength{\tabcolsep}{3pt}

\begin{minipage}[t]{0.48\linewidth}
\centering
\caption{Ablation studies of the impact of different spectra in SFDNet on AI-TOD dataset. The bold number indicates the best result obtained in the experiment.}
\label{tab:Spectrum_impact}
\begin{tabular}{ccc|cccc}
\hline
Low & Mid & High & AP$_\text{vt}$ & AP$_\text{t}$ & AP$_\text{s}$ & AP$_\text{m}$\\
\hline
$\times$ & $\checkmark$ & $\checkmark$ & 
11.4 & 24.4 & 29.4 & 35.0\\
$\checkmark$ & $\times$ & $\checkmark$ & 
10.0 & 23.4 & 27.3 & 33.2\\
$\checkmark$ & $\checkmark$ & $\times$ & 
9.2 & 21.2 & 28.5 & 34.8\\
$\checkmark$ & $\checkmark$ & $\checkmark$ & \textbf{14.4} & \textbf{29.0} & \textbf{33.4} & \textbf{40.8}\\
\hline
\end{tabular}
\end{minipage}
\hfill
\begin{minipage}[t]{0.48\linewidth}
\centering
\caption{Ablation studies of the spectrum aggregation mechanism in SFDNet on AI-TOD dataset. The bold number indicates the best result obtained in the experiment.}
\label{tab:spectrum_aggregation}
\begin{tabular}{c|ccccc}
\hline
Setting & AP & AP$_\text{vt}$ & AP$_\text{t}$ & AP$_\text{s}$ & AP$_\text{m}$\\
\hline
Attention & 26.6 & 12.3 & 27.0 & 30.9 & 38.3\\
VMamba & 26.4 & 12.9 & 26.9 & 30.3 & 36.5\\
Conv & 26.3 & 11.4 & 26.9 & 30.1 & 38.5\\
\textbf{S3M} & \textbf{29.0} & \textbf{14.4} & \textbf{29.0} & \textbf{33.4} & \textbf{40.8}\\
\hline
\end{tabular}
\end{minipage}
\end{table}

\noindent
\textbf{Hyperparameter Sensitivity Analysis.} ASD employs DoG-based spectral decomposition to disentangle features into low, mid, and high frequency components. As shown in Tab.~\ref{tab:sigma_k}, the performance is relatively insensitive to $\sigma$ and $k$. We attribute this robustness to the limited scale variation of small objects in the evaluated datasets, enabling ASD to generalize well across different input sizes without extensive hyperparameter tuning.
\begin{table}[!t]
    \centering
    \caption{Ablation studies on parameters $k$ and $\sigma$.}
    \label{tab:sigma_k}
    \setlength{\tabcolsep}{8pt}
        \begin{tabular}{l|cccc}
        \hline
        $k$ & 1.2 & 1.4 & 1.6 & 1.8 \\
        \hline
        mAP & 28.8 & 29.0 & 29.1 & 28.9 \\
        \hline
        \end{tabular}%
        \quad 
        \begin{tabular}{l|cccc}
        \hline
        $\sigma$ & 0.5 & 1.0 & 1.5 & 2.0 \\
        \hline
        mAP & 28.9 & 29.0 & 28.8 & 28.7 \\
        \hline
        \end{tabular}%
\end{table}

\subsection{Qualitative Visualization}
\begin{figure*}[!t]
\centering
\caption{Visualization of feature maps at the P2 level on SODA-D dataset. The first row presents the results of DNTR, while the second row shows those of SFDNet.}
\includegraphics[width=1.0\linewidth]{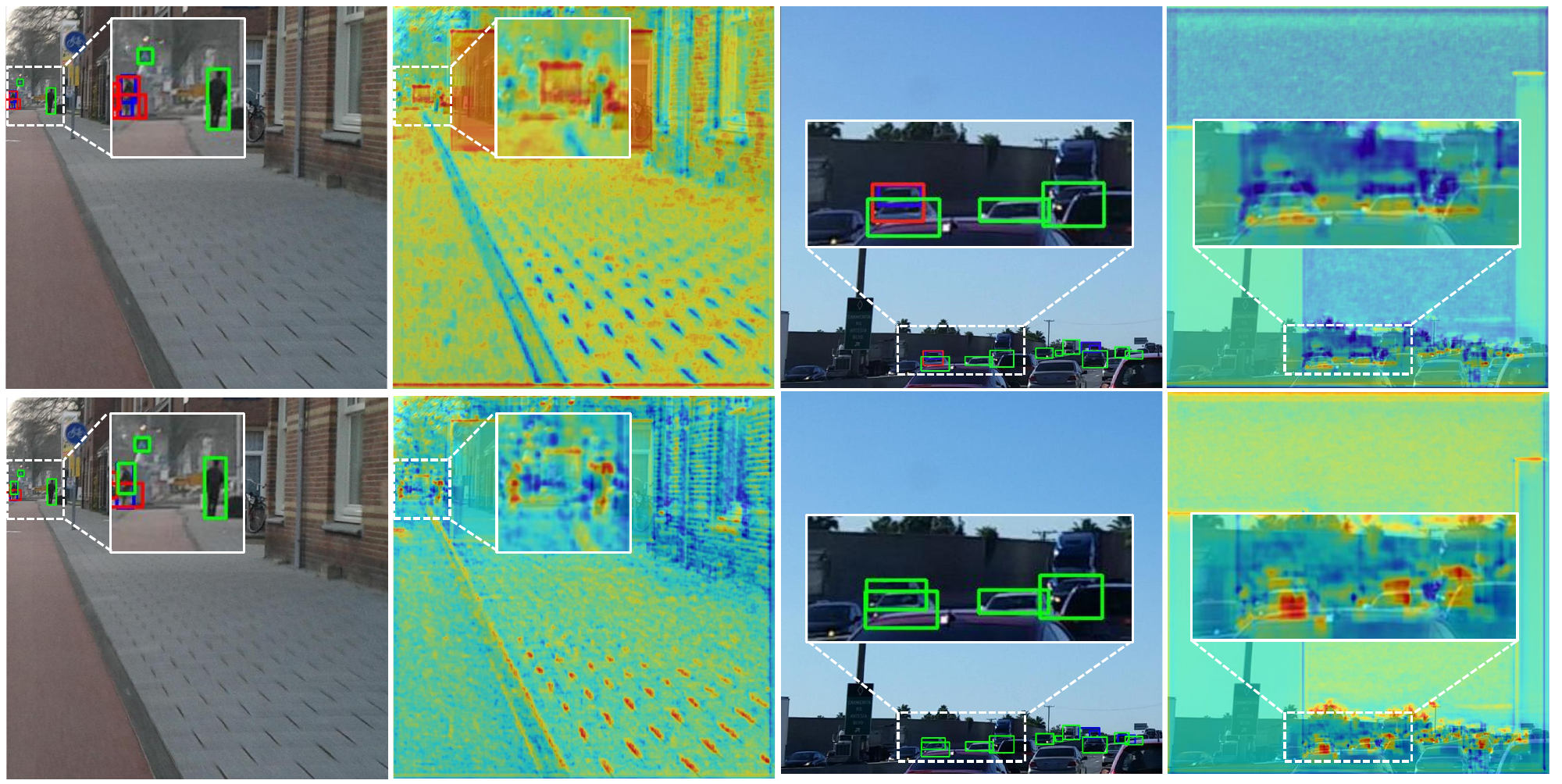}
\label{fig:heatmap_visualization_aitod}
\end{figure*}

\noindent\textbf{Heatmap Visualization.} To qualitatively assess the effectiveness of our method, we visualize the feature heatmaps at the P2 level. As shown in Fig.\ref{fig:heatmap_visualization_aitod}, the first row shows the results of DNTR, while the second row presents those of SFDNet. Compared with DNTR, SFDNet demonstrates stronger discriminative capability in its feature representations.

\noindent\textbf{Qualitative Analysis of Per-Class Performance.} 
We visualize per-class performance on the SODA-D and AI-TOD datasets. As shown in Fig.~\ref{fig:visualization_radar}, our method achieves the best performance across all classes, validating the effectiveness of the proposed Adaptive Spectrum Disentanglement (ASD) module and Class-Wise Prototype Distillation (CPD) procedure. 

\begin{figure*}[!t]
\centering
\caption{Visualization of SFDNet feature maps for different spectral components at the P2 level on the AI-TOD test set. From left to right: ground-truth (GT), high, mid, and low frequency spectra.}
\includegraphics[width=1.0\linewidth]{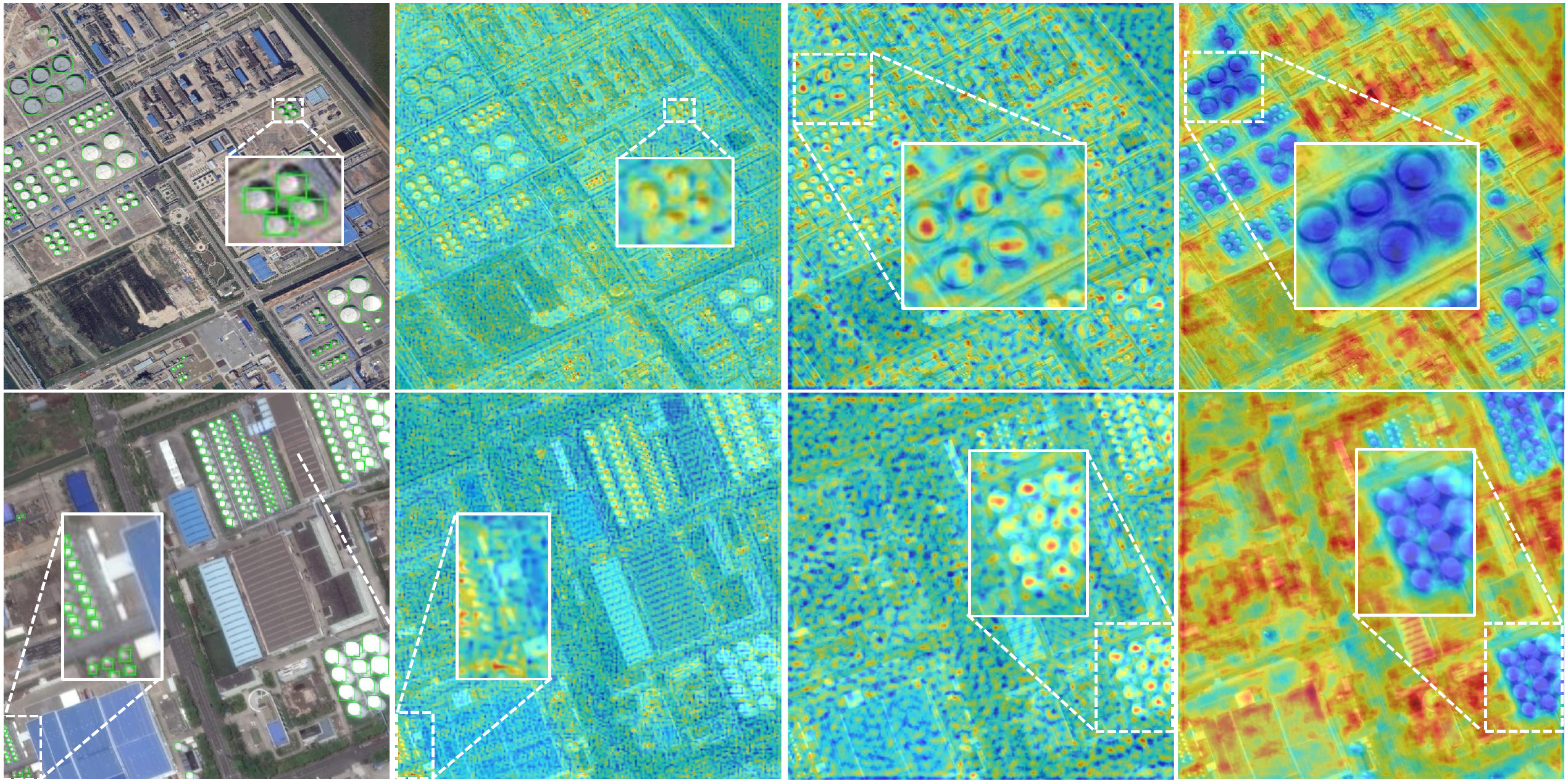}
\label{fig:heatmap_visualization_diff_spectra}
\end{figure*}

\noindent\textbf{Heatmap of Different Spectral Components.} Fig.~\ref{fig:heatmap_visualization_diff_spectra} visualizes the heatmaps of different spectral components. It can be observed that the low-frequency spectrum exhibits weak responses in the target regions, the mid-frequency spectrum responds strongly to relatively large objects (e.g., the storage tank in the zoomed-in region), and the high-frequency spectrum preferentially attends to smaller objects. These observations demonstrate the effectiveness of our spectrum disentanglement mechanism.

\section{Conclusion}
In this paper, we propose a novel small object detection framework, termed \textbf{SFDNet}, which effectively enhances the detection of small objects via spectrum-aware feature disentanglement. Specifically, SFDNet decomposes features into multiple spectral components and employs spectrum-specific scanning strategies to capture the key semantic features of small targets within each spectral component, thereby achieving full-spectrum noise suppression and yielding discriminative feature representations. In addition, we design a Class-Wise Prototype Distillation (CPD) procedure that enforces consistent feature representations within each class. Extensive experiments on public benchmarks demonstrate that SFDNet achieves state-of-the-art performance, validating the effectiveness of the proposed method.

\noindent
\textbf{Acknowledgment.}
This work is supported by the National Natural Science Foundation of China (No. 62402055), Beijing Natural Science Foundation (Grant No.L2603037).


%
%





\end{document}